\documentclass[11pt]{article}

\usepackage[final]{acl}

\usepackage{latexsym}
\usepackage[T1]{fontenc}
\usepackage[utf8]{inputenc}
\usepackage{microtype}
\usepackage{inconsolata}
\usepackage{graphicx}
\usepackage{booktabs}
\usepackage{multirow}
\usepackage{amsmath}
\usepackage{float}
\usepackage{hyperref}

\usepackage{fontspec}
\newfontfamily\devanagari[Script=Devanagari]{NotoSansDevanagari.ttf}
\newfontfamily\telugu[Script=Telugu]{NotoSansTelugu.ttf}
\newfontfamily\kannada[Script=Kannada]{NotoSansKannada.ttf}
\newfontfamily\cyrillicfont[Script=Cyrillic,ItalicFont=NotoSans-Italic.ttf]{NotoSans-Regular.ttf}

\title{The Invisible Gap: Romanized Indic Generation Fails Where Automatic Metrics Succeed}

\author{%
\begin{tabular}{cc}
Surbhi Sharma\textsuperscript{\textdagger} &
Devesh Maheshwari\textsuperscript{\textdagger} \\
Houston Methodist & University of Wisconsin--Madison \\
\texttt{ssharma3@houstonmethodist.org} &
\texttt{dmaheshwar22@wisc.edu} \\[0.8em]
Lokesh Meesala\textsuperscript{\textdagger} &
Madhu Sudhan Rao Beesetty\textsuperscript{\textdagger} \\
Morgan Stanley & PayPal \\
\texttt{lmeesala@students.kennesaw.edu} &
\texttt{madhusudhanbeesetty@gmail.com}
\end{tabular}
}

\date{}

\begin{document}

\maketitle
\begingroup
\renewcommand{\thefootnote}{\fnsymbol{footnote}}
\footnotetext[2]{All authors contributed equally and share first authorship. Author order does not indicate relative contribution.}
\endgroup

\begin{abstract}
Hundreds of millions of Indians communicate with digital systems in romanized Latin characters---typing \textit{aap kaise hain} ({\devanagari आप कैसे हैं}) rather than native script---yet romanized Indic generation is absent from NLP benchmarks and evaluation frameworks. We demonstrate that automatic metrics (BGE-M3, BERTScore) systematically misrepresent the quality of romanized Hindi, Telugu, and Kannada generation. We evaluate four LLMs (Claude Opus 4.8, GPT-5.5, Gemini 2.5-flash, Qwen3.5-35B) on 142 OpenAssistant query-response pairs. On a 72-response diagnostic subset, native speakers rate GPT-5.5 1.43 points higher than Claude Opus 4.8 on a 12-point rubric, whereas their full-dataset BGE-M3 means differ by only 0.004. Two distinct model-specific failure modes emerge: Gemini produces script spillage (16.9\% of Telugu responses leak native script), while Qwen collapses to English-only output (16.2\% of Kannada responses). We introduce a failure-aware evaluation policy that assigns zero to empty outputs, script-corrupted responses, and instruction collapses---revealing failures metrics alone cannot detect. Our results show that evaluating romanized Indic generation requires a coordinated bundle: semantic similarity, script compliance, failure coverage, and native-speaker acceptability.
\end{abstract}

\section{Introduction}
\subsection{The Romanization Reality}

For millions of Indian language speakers, the practical interface to digital systems is not native script but romanized Latin characters. A Hindi speaker types \textit{aap kaise hain} ({\devanagari आप कैसे हैं}, ``how are you?''); a Telugu speaker types \textit{meeru ela unnaru} ({\telugu మీరు ఎలా ఉన్నారు}); a Kannada speaker types \textit{neevu hegiddira} ({\kannada ನೀವು ಹೇಗಿದ್ದೀರ}). Although these sentences use Latin letters, they are not English translations. Their vocabulary, grammar, and social register belong entirely to the native language. Romanized text is therefore not a mere transliteration convenience for users without keyboard setup---it is the lived interface through which hundreds of millions of people access digital services, including large language models.

This behavior is quantified in real-world data. A corpus of patient WhatsApp interactions shows romanization accounts for 83\% of Hindi messages and 73\% of Telugu messages~\cite{khullar2025scriptgap}. The Dakshina corpus~\cite{roark2020dakshina} and user studies confirm this dominance across languages. Yet most multilingual NLP benchmarks evaluate only formal native-script text: XQuAD~\cite{artetxe2020xquad}, IndicQA~\cite{doddapaneni2023indicqa}, and MLQA~\cite{lewis-etal-2020-mlqa} contain no romanized data. Datasets like Dakshina document romanization \emph{variability} but target transliteration, not generation quality. Consequently, the ability of modern LLMs to generate long, natural, semantically faithful romanized Indic text remains fundamentally uncharacterized.

\subsection{The Metric Blindness Problem}

When LLM outputs are scored automatically, metrics like BERTScore and BGE-M3 operate on embedding similarity or token overlap. For open-ended generation with no ground-truth reference, these metrics cannot distinguish a genuine romanized response from three distinct failures: a verbatim English copy (word-overlap ratio $=1.0$), which is perfectly similar to the English reference but a complete task failure; a response written in native script, which preserves meaning but violates the interface the user expects; and a response using English syntactic structure with Indic words substituted in, which passes embedding similarity but reads as translated rather than naturally composed. Standard metrics are blind to these failures because they measure \emph{similarity to an English reference}, not \emph{acceptability in romanized Indic}. The gap between what metrics score and what users need is therefore invisible to automated systems; only native speakers, evaluating the output directly, can catch it.

\subsection{Contributions}

This paper makes four contributions:

\begin{enumerate}
    \item \textbf{Quantified metric-human divergence:} We measure a 1.43-point native-speaker quality gap (GPT-5.5 10.73 vs.\ Claude 9.30 out of 12) that automatic metrics compress to 0.004---a 357$\times$ underestimation.

    \item \textbf{Model-specific failure taxonomy:} We identify two non-overlapping failure modes: Gemini 2.5-flash exhibits script spillage (up to 16.9\% of Telugu responses), while Qwen3.5-35B exhibits instruction collapse (up to 16.2\% of Kannada responses).

    \item \textbf{Failure-aware evaluation methodology:} We introduce a policy that assigns zero to invalid rows (empty, English-only, script-corrupted, or degenerate) rather than excluding them, ensuring all 142 samples enter aggregate means and revealing reliability alongside semantic quality.

    \item \textbf{Reusable native-speaker rubric:} We provide a 10-dimension evaluation rubric (naturalness, fluency, register, cultural fit, script purity, overall usability) that can be reused for future romanized Indic generation studies.
\end{enumerate}

\subsection{Scope and Design}

We test four multilingual LLMs representing both proprietary APIs and open-source models: Claude Opus 4.8 (Anthropic), GPT-5.5 (OpenAI), Gemini 2.5-flash (Google), and Qwen3.5-35B (Alibaba, served locally). We evaluate on three Indic languages---Hindi, Telugu, Kannada---using 142 high-quality English query-response pairs from OpenAssistant~\cite{kopf2023openassistant}. Each model receives the same inputs and returns romanized outputs under a single structured prompt, enabling direct cross-model and cross-language comparison.

Evaluation combines two complementary approaches: (1) automatic metrics including BERTScore~\cite{zhang2020bertscore} and BGE-M3 cosine similarity~\cite{chen2024bgem3} applied under a failure-aware policy, and (2) native-speaker judgment on a strategically sampled diagnostic set (24 responses per language, 72 total). This design directly measures the gap between what metrics report and what native speakers require.

\section{Related Work}
The starting point for this work is a tokenization inequality. Petrov et al.~\cite{petrov2023tokenization} showed that tokenizers trained on English-dominant corpora require up to 15$\times$ more tokens to represent the same content in non-English languages, creating a structural cost asymmetry. Prior work on romanized input has begun to characterize model quality under this constraint. Khullar et al.~\cite{khullar2025scriptgap} showed romanized input degrades LLM performance on structured classification tasks by up to 24 points relative to native script. RomanSetu~\cite{husain2024romansetu} took a complementary view---using romanization as an \emph{input strategy} to unlock multilingual capabilities---and found improvements on formal benchmarks. Both studies evaluate tasks with a known correct answer.

\textbf{The evaluation gap for open-ended generation remains unaddressed.} Standard benchmarks such as XQuAD~\cite{artetxe2020xquad}, IndicQA~\cite{doddapaneni2023indicqa}, and MLQA~\cite{lewis-etal-2020-mlqa} presuppose a reference answer and use formal native script. Corpora like Dakshina~\cite{roark2020dakshina} and Aksharantar~\cite{madhani2023aksharantar} document romanization \emph{variability} but target transliteration (converting script, not generating new text). Multilingual BERTScore~\cite{zhang2020bertscore} and BGE-M3~\cite{chen2024bgem3} compare outputs to a reference but have no mechanism to detect wrong script (native characters appearing mid-romanized text), unnatural register (text that sounds like word-by-word translation), or instruction collapse (returning English unchanged when romanization was requested).

Beyond romanization, the broader literature on LLM-as-judge evaluation raises a further concern: judges exhibit systematic biases, favoring verbose or stylistically confident responses over substantively correct ones~\cite{zheng2023judging}. This motivated our use of native-speaker ground truth rather than LLM calibration.

We fill this gap by using native-speaker human evaluation as ground truth and quantifying the divergence from automatic metrics on a task (open-ended generation in romanized Indic) where neither prior romanization studies nor standard multilingual evaluation frameworks offer empirical evidence.

\section{Methodology}
\label{sec:method}
\subsection{Data}

We use 142 English query-response pairs derived from OpenAssistant (OASST2), a high-quality conversation dataset. Each pair consists of a user query and a model-generated response. We selected pairs by filtering for English-language assistant turns and stratifying across prompt types: technical (how-to, debugging), emotional (advice, support), narrative (storytelling, creative), and factual (definitions, explanations). This diversity ensures the evaluation covers real-world use patterns. The same 142 pairs are used for all three target languages and all four models, enabling direct comparison under identical conditions.

\subsection{Models}

We evaluate four LLMs spanning proprietary APIs and open-source deployments, all at temperature 0: \textbf{Claude Opus 4.8} (Anthropic API, up to 4096 tokens), \textbf{GPT-5.5} (OpenAI API), \textbf{Gemini 2.5-flash} (Google Generative AI API), and \textbf{Qwen3.5-35B} (Alibaba), the last served locally via Ollama on an NVIDIA RTX PRO 6000 Blackwell GPU with up to 4096 tokens.

\subsection{Generation Protocol}

Each model receives a structured prompt instructing it to return a JSON object with two fields: \texttt{romanized\_query} and \texttt{romanized\_response}. The prompt specifies Latin-script output only (no native script) and requires natural English code-mixing where it improves readability. Kannada receives a stricter variant prohibiting both native-script leakage and English structural borrowing, reflecting the language's distinct phonetic system.

Joint generation (both query and response in one call) is used rather than independent generation. Pilot experiments showed independent generation caused inconsistencies: the model would romanize the query one way in isolation, then produce inconsistent romanization in the response. Jointly generating both fields preserves conversational context and ensures coherence.

Malformed JSON outputs are retained when recoverable (e.g., missing quotes); failed fields remain blank. This strategy avoids discarding data and allows the failure analysis to include partial outputs.

\subsection{Automatic Evaluation Metrics}

\textbf{BGE-M3 cosine similarity.} BGE-M3 is a dense embedding model trained on multilingual corpora. For each non-empty response, we compute the cosine similarity between the original English response and its romanized counterpart in BGE-M3's dense vector space. It is our primary cross-lingual semantic measure because it requires no parallel reference and handles multilingual content natively.

\textbf{BERTScore F1.} BERTScore measures token-level alignment using contextual embeddings from \texttt{bert-base-multilingual-cased} (without baseline rescaling). We report F1 as a secondary signal; its token-alignment perspective complements BGE-M3's dense semantic measure.

\textbf{Failure-aware coverage policy.} Standard evaluation excludes invalid outputs; we instead assign them zero and include them in aggregate means, surfacing failures that metrics mask. A row is marked invalid if any of four conditions hold: (i) an empty or whitespace-only response; (ii) an English-only copy identical to the source (word-overlap ratio $=1.0$), a complete task failure; (iii) script spillage---one or more native-script characters, with legitimate romanization diacritics (\={a}, \={e}, \d{t}) excluded; or (iv) degenerate output---a single token repeated $\geq 20$ times, or a four-token phrase repeated $\geq 50$ times. All invalid rows receive zero on every automatic metric and remain in the denominator, so reliability and semantic quality are measured on the same 142-sample base.

\subsection{Native-Speaker Human Evaluation}
\label{sec:human-eval-method}

\subsubsection{Sampling Strategy}

We selected a diagnostic set of 72 responses (24 per language) using stratified sampling across six categories: pure or mostly English outputs that metrics still score highly; native-script leakage (including cross-script contamination, e.g., Telugu characters in a Kannada response); metric-divergence rows where BGE-M3 and BERTScore disagree by more than 0.15; hard creative, emotional, or roleplay prompts; clean high-quality anchors for calibration; and English-task rows (OCR cleanup, grammar correction, email improvement) where English output is correct and which are scored separately. Because rows are selected on failure conditions, per-model counts within this set do not represent population failure rates; population statistics are computed separately over all 142 responses per model.

\subsubsection{Rubric Design}

Native speakers scored each response on a rubric covering response completeness, script purity, English--romanization balance, romanization consistency, semantic faithfulness, naturalness and fluency, grammar correctness, word choice accuracy, register appropriateness, and overall usability. Full per-dimension definitions, point allocations, and scoring examples are provided in Appendix~\ref{sec:rubric}.

\subsubsection{Evaluation Process}

Three native speakers (one per language: Hindi, Telugu, Kannada) independently scored the 72 diagnostic rows. A single LLM-generated reference score (Claude Opus 4.8) was provided only as a calibration anchor; evaluators scored every response from scratch without reference to it. This design isolates native-speaker judgment from LLM influence.

Formal inter-rater agreement was not measured due to single-rater-per-language design (necessary to maintain language fluency and cultural authenticity). However, findings from the three independent evaluators show consistent patterns, suggesting rubric clarity.

\subsection{Code Availability}

The code and supporting resources are publicly available at \url{https://github.com/MadhuBeesetty/llm-romanization}.

\section{Results}
\label{sec:results}
\subsection{Automatic Evaluation}

Table~\ref{tab:main-results} reports failure-aware BGE-M3 cosine similarity and BERTScore F1 for all four models across the three languages. All 142 samples enter each mean; rows flagged as invalid (empty output, English-only copy, native-script corruption, or repetition loop) receive zero for all metrics rather than being excluded.

\begin{table}[t]
\centering
\footnotesize
\setlength{\tabcolsep}{4pt}
\begin{tabular}{llccc}
\toprule
Model & Lang. & BGE & BERT & Inv. \\
\midrule
\multirow{3}{*}{Claude Opus-4.8} & Hindi   & 0.837 & 0.676 & 0 \\
                            & Telugu  & 0.789 & 0.671 & 4 \\
                            & Kannada & 0.733 & 0.620 & 5 \\
\midrule
\multirow{3}{*}{GPT-5.5}    & Hindi   & 0.828 & 0.663 & 0 \\
                            & Telugu  & 0.794 & 0.667 & 0 \\
                            & Kannada & 0.749 & 0.641 & 2 \\
\midrule
\multirow{3}{*}{Gemini-2.5-flash}    & Hindi   & 0.763 & 0.617 & 10 \\
                            & Telugu  & 0.630 & 0.548 & 25 \\
                            & Kannada & 0.664 & 0.569 & 19 \\
\midrule
\multirow{3}{*}{Qwen3.5-35B}    & Hindi   & 0.787 & 0.639 & 4  \\
                            & Telugu  & 0.598 & 0.574 & 16 \\
                            & Kannada & 0.537 & 0.549 & 20 \\
\bottomrule
\end{tabular}
\caption{Failure-aware automatic evaluation scores (response only). BGE is BGE-M3 cosine similarity; BERT is BERTScore F1 using \texttt{bert-base-multilingual-cased}. Inv. counts rows receiving zero (empty, English-only, script corruption, repetition) out of 142 total.}
\label{tab:main-results}
\end{table}

\begin{figure}[t]
  \centering
  \includegraphics[width=\linewidth]{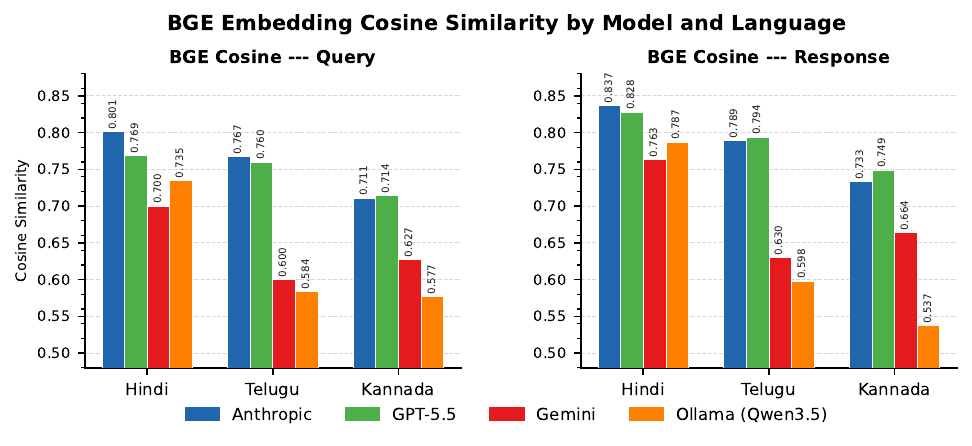}
  \caption{BGE-M3 cosine similarity for romanized queries (left) and responses (right) across all four models and three languages. Hindi shows highest scores (0.763--0.837), with Telugu and Kannada showing lower scores due to reduced training data availability.}
  \label{fig:bge-cosine}
\end{figure}

\begin{figure}[t]
  \centering
  \includegraphics[width=\linewidth]{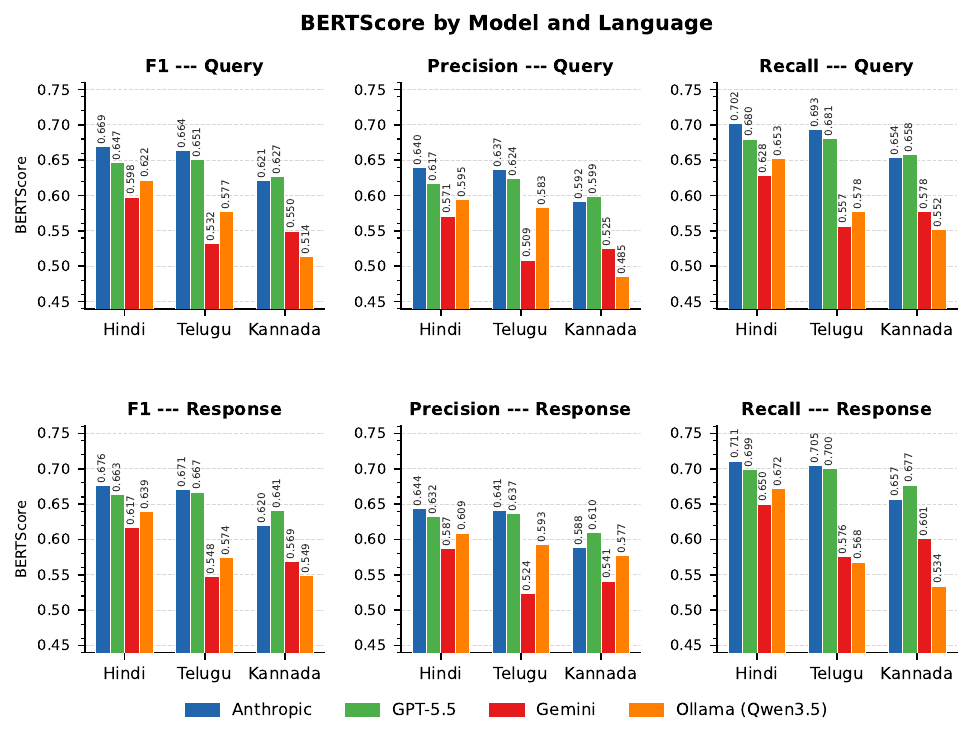}
  \caption{Multilingual BERTScore (F1, Precision, Recall) for romanized queries (top row) and responses (bottom row) across all four models and three languages. Scores use \texttt{bert-base-multilingual-cased} and are failure-aware (invalid rows receive zero).}
  \label{fig:bert-scores}
\end{figure}

Across languages, Hindi obtains the highest scores across all four models (BGE: 0.763--0.837), followed by Telugu (0.598--0.794) and Kannada (0.537--0.749). This ordering is consistent with the availability of romanized Hindi training data: Hinglish is the most widely represented romanized Indic variety in publicly available corpora, while romanized Kannada and Telugu appear far less frequently.

At the model level, Claude Opus 4.8 and GPT-5.5 produce the strongest and most reliable outputs. Both achieve zero invalid rows for Hindi, and at most two for Telugu and Kannada. Their BGE scores for Hindi (0.828--0.837) are the highest observed. Gemini 2.5-flash and Qwen3.5-35B degrade sharply on Telugu and Kannada, but for distinct reasons examined in Section~\ref{sec:failures}.

Despite these rankings, the BGE score range for the two strongest models is misleadingly narrow: Claude Opus 4.8 and GPT-5.5 produce scores between 0.733 and 0.837 across all language conditions, a band that appears to indicate acceptable semantic preservation in isolation. Section~\ref{sec:human-eval} shows that native speakers do not agree. Figures~\ref{fig:bge-cosine} and~\ref{fig:bert-scores} visualize the full BGE-M3 and BERTScore distributions across all model--language combinations.

\subsection{Failure Analysis}
\label{sec:failures}

Table~\ref{tab:failures} breaks down the two primary failure modes: responses identical to the English source (repetition) and responses containing native script characters (script corruption).

\begin{table}[t]
\centering
\footnotesize
\setlength{\tabcolsep}{4.5pt}
\begin{tabular}{llrr}
\toprule
Model & Lang. & Eng.-only & Script sp. \\
\midrule
\multirow{3}{*}{Claude 4.8} & Hindi   & 0  & 0  \\
                            & Telugu  & 0  & 3  \\
                            & Kannada & 0  & 5  \\
\midrule
\multirow{3}{*}{GPT-5.5}    & Hindi   & 0  & 0  \\
                            & Telugu  & 0  & 0  \\
                            & Kannada & 0  & 2  \\
\midrule
\multirow{3}{*}{Gemini-2.5} & Hindi   & 0  & 8  \\
                            & Telugu  & 0  & 24 \\
                            & Kannada & 1  & 19 \\
\midrule
\multirow{3}{*}{Qwen-3.5}   & Hindi   & 3  & 0  \\
                            & Telugu  & 16 & 0  \\
                            & Kannada & 19 & 0  \\
\bottomrule
\end{tabular}
\caption{Failure breakdown per model and language (out of 142). Eng.-only: response returned in English (word-overlap $=1.0$, strict match to source) instead of the romanized target. Script sp.\ (spillage): response contains one or more non-Latin characters (native script, Cyrillic, etc.), after excluding legitimate romanization diacritics (\={a}, \={e}, \d{t}). A broader threshold (word-overlap $\geq 0.90$) yields 70 English-only cases across all models; the strict $=1.0$ threshold used here counts only verbatim copies (54 cases).}
\label{tab:failures}
\end{table}

Gemini 2.5-flash and Qwen3.5-35B represent qualitatively different failure modes. Gemini produces \emph{script spillage}: non-Latin characters appear mid-sentence in otherwise romanized output. A detailed character-level scan of all responses (after excluding legitimate romanization diacritics such as \={a}, \={e}, \d{t} and a Catalan source-title character in one play prompt) reveals that spillage affects 24 of 142 Telugu responses (16.9\%), 19 of 142 Kannada responses (13.4\%), and 8 of 142 Hindi responses (5.6\%) from Gemini. The Telugu case is the most severe: Gemini 2.5-flash produces entire paragraphs in native Telugu script across 24 rows, totaling 2,635 Telugu characters. Claude Opus 4.8 shows minor spillage confined to Kannada (5 rows) and Telugu (3 rows). GPT-5.5 is nearly clean, with only 21 Kannada characters across 2 rows. Qwen3.5-35B produces zero script spillage across all three languages.

These patterns reveal diverse failure sub-types (including cross-script contamination and unexpected character insertion) that confirm Gemini 2.5-flash's script boundary is unstable in ways BGE-M3 cannot detect.

Qwen3.5-35B's dominant failure is \emph{English-only output}: the model returns the answer in English rather than romanizing it. Of 70 English-only instances detected across all models ($\geq$0.90 word-overlap), Qwen3.5-35B accounts for 51, of which 41 are genuine failures on informational and conversational prompts; GPT-5.5 produces zero. English-only rates for Qwen3.5-35B reach 11.3\% for Telugu and 16.2\% for Kannada, while Hindi is largely unaffected at 2.8\%.

A second Qwen3.5-35B-specific failure is \emph{degeneration}: the model collapses into repeating tokens, generating the same word or phrase hundreds of times until hitting token limits. We identify 13 such instances---every one from Qwen3.5-35B, zero from other models. Human raters assign usability zero to every degeneration case, while BGE-M3 assigns scores of 0.38--0.72, sometimes still ``passing.''

The two failure types do not overlap: Gemini 2.5-flash never produces English-only output and Qwen3.5-35B never leaks native script, suggesting the failures originate in different aspects of instruction following. BERTScore and BGE-M3 treat both failure types under the failure-aware policy: they receive zero. Without this policy---if invalid rows were silently excluded---Gemini 2.5-flash's Telugu BGE score would rise from 0.630 to approximately 0.82, indistinguishable from GPT-5.5's 0.794. The failure-aware denominator is therefore not cosmetic: it materially changes model ranking.

The correlation between automatic scores and English word overlap directly illustrates why these metrics miss task failures. On the Qwen3.5-35B outputs, raw BERTScore-F1 correlates with English word-overlap at $r = 0.955$ and BGE-M3 at $r = 0.805$: both metrics largely reward how much of the output remains English rather than how well it is romanized. Consistent with this, Qwen3.5-35B's verbatim-English Telugu responses receive BGE $= 1.000$ and BERTScore-F1 $= 1.000$, while its genuine romanized Telugu averages only $0.67$---the metric scores a complete task failure higher than a success. Raw BERTScore-F1 is further compressed by a high floor: empty responses still score ${\approx}0.42$, because baseline rescaling is disabled and multilingual-BERT embeddings occupy a narrow cone. These properties explain why the metric assigns its highest scores to its worst failures.

\subsection{Human Evaluation}
\label{sec:human-eval}

This BGE band conceals quality failures that only native speakers detect: wrong register, unnatural word choice, and English-patterned syntax.

Native speakers rated 72 responses (24 per language) on five dimensions, max 12 points (Section~\ref{sec:method}). An LLM reference score showed poor agreement with native judgments (MAD $\approx$4/12): one Gemini 2.5-flash response scored 10.4/12 by the LLM but 3/12 by the native reader. All reported scores are native-speaker judgments; LLM substitutability is discussed in Section~\ref{sec:discussion}.

\begin{table*}[t]
\centering
\small
\setlength{\tabcolsep}{5pt}
\begin{tabular}{lccccccccc}
\toprule
Model & Nat. & Gram. & Word & Reg. & Usab. & Total & BGE-M3 & BERT-F1 & Eng.-only \\
      & /3   & /2    & /2   & /2   & /3    & /12   & (mean) & (mean)  & fails \\
\midrule
GPT-5.5         & 2.49 & 1.93 & 1.97 & 1.94 & 2.40 & \textbf{10.73} & \textbf{0.790} & \textbf{0.657} & 0  \\
Claude Opus 4.8 & 2.05 & 1.66 & 1.70 & 1.73 & 2.15 & 9.30           & 0.786          & 0.656          & 2  \\
Gemini 2.5-flash & 1.89 & 1.73 & 1.54 & 1.70 & 2.06 & 8.91          & 0.686          & 0.578          & 2  \\
Qwen3.5-35B     & 0.76 & 0.89 & 0.93 & 1.04 & 0.85 & 4.48           & 0.641          & 0.587          & 41 \\
\bottomrule
\end{tabular}
\caption{Human evaluation with automatic-metric summary. Per-dimension human scores are native-speaker judgments (weighted mean over 72 rows, 24 per language): Nat.=Naturalness/Fluency, Gram.=Grammar, Word=Word Choice, Reg.=Register, Usab.=Usability; Total is their sum (max 12). BGE-M3 and BERT-F1 are failure-aware means over all 142 responses per model. Eng.-only fails: genuine cases (excluding English-task prompts) where the model returned English instead of the romanized target.}
\label{tab:human-eval}
\end{table*}

Table~\ref{tab:human-eval} exposes the central finding of this paper: \textbf{automatic metrics are dangerously lenient on per-instance failures}. BGE-M3 scores for Claude Opus 4.8 (0.786) and GPT-5.5 (0.790) are nearly indistinguishable---a difference of 0.004 points that no downstream decision would act on. Native speakers disagree sharply: GPT-5.5 scores 10.73 out of 12 while Claude Opus 4.8 scores 9.30, a gap of 1.43 points driven primarily by naturalness (2.49 vs 2.05) and register appropriateness (1.94 vs 1.73). Both metrics and humans identify GPT-5.5 as the stronger model, but metrics assign false confidence in near-equivalence where native speakers observe meaningful quality difference.

The gap is most severe for Qwen3.5-35B. BGE-M3 assigns it 0.641---poor, but not catastrophic by the metric's scale. Human raters assign a mean of 4.48 out of 12, with Telugu and Kannada responses averaging 4.07 and 2.86 respectively (largely unusable), while Hindi averages 8.28---a divergence explained by Qwen3.5-35B's near-complete instruction collapse in Dravidian languages (30\% of Kannada outputs returned in English) versus better Hindi coverage in its training data. BGE cannot distinguish between a response that is semantically adjacent but linguistically broken and one that is simply the English source returned verbatim---both score near 1.000 under that policy. Human raters make this distinction immediately.

Hindi obtains the highest human mean (9.42/12). Notably, the Telugu/Kannada ordering reverses relative to automatic metrics: human raters score Kannada (7.11/12) above Telugu (6.89/12), suggesting that Kannada outputs, while fewer in number and harder for models to generate, receive higher quality marks from native speakers when they succeed. Register appropriateness is the most commonly flagged failure dimension in evaluator notes: Gemini 2.5-flash responses in all three languages were described as ``too formal'' or using ``very formal words'' inconsistent with the conversational register of the prompts. One GPT-5.5 Kannada response was noted as ``almost English''---a high-overlap output that BGE would score near 1.0 but that native speakers identified as a task failure.

\section{Analysis}
\label{sec:analysis}
\subsection{Why Metrics Mask Failures}

As Section~\ref{sec:results} shows, BGE-M3 scores for the two strongest models cluster narrowly (0.73--0.84), appearing to indicate acceptable semantic preservation. This appearance is misleading because metrics systematically assign high confidence to per-instance failures: English verbatim copies score BGE $=1.000$ despite being complete task failures, and responses containing full paragraphs of native script still score 0.80--0.84 because semantic content is preserved even as the Latin-script interface constraint is violated. In both cases native speakers reject outputs that automatic metrics reward with near-perfect scores. This is not a ranking problem---metrics correctly identify the stronger models---but a confidence problem: metrics provide false assurance of safety on outputs that should trigger alerts.

The metrics are moreover unreliable relative to each other: BGE-M3 and BERTScore diverge by more than 0.15 on 44\% of rows, reaching 0.28 on a single response. When two standard metrics disagree on nearly half of all outputs, neither can be trusted as ground truth for this task, and their agreement should not be interpreted as validation.

\subsection{Cross-Script Contamination and Degeneration}

Beyond the aggregate spillage and English-only rates reported in Section~\ref{sec:failures}, the qualitative character of the failures is itself diagnostic. Gemini 2.5-flash's script spillage is not confined to the target native script: Telugu row~53 inserts the Russian word \textit{{\cyrillicfont единственный}} (romanized: \textit{yedinstvenniy}, ``the only one'') mid-sentence; Kannada row~33 emits Telugu script ({\telugu అవును}) rather than Kannada romanization; Telugu row~19 mixes Tamil and Telugu in a single line; and eight Hindi rows recur the Devanagari token {\devanagari मानना} (romanized: \textit{manna}, ``to believe''). This cross-script leakage indicates an unstable script-selection boundary, not a simple romanization shortfall---a distinction no embedding metric can register.

Qwen3.5-35B exhibits a second, orthogonal instability. Its English-only fallback is concentrated in the Dravidian languages (Section~\ref{sec:failures}), pointing to insufficient romanized Telugu and Kannada training data: when the model cannot romanize, it either returns English or collapses into repetition loops that run until the token limit. Kannada is worst affected, with 49 rows hitting the 4{,}096-token limit, which raises its mean output-token usage from 541 to 1{,}768 (Table~\ref{tab:output-tokens}).

\subsection{Register and Naturalness}

Independent analysis of romanized outputs reveals a consistent register problem that no automatic metric captures. All three API models (Claude, GPT, Gemini) generate text that follows English syntactic structure with Indic words inserted at corresponding positions---a pattern native speakers immediately identify as translated rather than natively composed.

Gemini 2.5-flash Hindi exhibits this most starkly, consistently using hyper-formal Sanskrit-register vocabulary ({\devanagari सांस्कृतिक महत्व} \textit{saanskritik mahatva}, {\devanagari अधिकार होता है} \textit{adhikar hota hai}) that no casual speaker would use in digital communication. GPT-5.5 Kannada uses more idiomatic affirmatives ({\kannada ಹೌದು} \textit{Haudu}, {\kannada ಖಂಡಿತ} \textit{Khandita}), signaling greater lexical competence, but collapses into the same translated-English architecture for longer expository prose.

This naturalness gap is captured by the native-speaker rubric but invisible to embedding-based metrics. Claude Opus 4.8 shows the most natural-sounding prose, with fewer instances of literal-translation artifacts.

\section{Discussion}
\label{sec:discussion}
\subsection{What the Gap Tells Us}

The core issue is not that metrics rank models wrong---they correctly identify GPT-5.5 $\geq$ Claude $\geq$ Gemini $\geq$ Qwen---but that they assign false confidence in equivalence where native speakers observe meaningful differences (Section~\ref{sec:human-eval}). For deployment, this is dangerous: a model scorer might conclude two models are interchangeable based on a 0.004-point metric gap, when native speakers would reject one significantly more often than the other.

The non-overlap of the two failure modes---Gemini never copies English, Qwen never leaks script---suggests they arise from different causes: Gemini struggles with script selection (which language/script to output), while Qwen struggles with language capacity (insufficient romanized-Indic exposure leads to task collapse). This distinction could inform future model tuning, and the failure taxonomy itself is robust regardless of rater count.

\subsection{Metric Reliability}

Neither automatic metric is sufficient alone: BGE-M3 and BERTScore disagree by more than 0.15 on 44\% of rows (Section~\ref{sec:analysis}), and each captures phenomena the other misses. For romanized generation evaluation, combining multiple metrics with human judgment is essential.

\section{Conclusion}
Hundreds of millions of Indians communicate with AI systems in romanized Latin script, yet the quality of romanized generation remains invisible to standard evaluation metrics. This paper quantifies the gap: native speakers detect quality failures that BGE-M3 and BERTScore systematically miss, with a 1.43-point human difference compressed to 0.004 by metrics.

Two distinct model-specific failure modes emerged---Gemini's script spillage and Qwen's instruction collapse---neither of which co-occur in the same model. These failures change model rankings when included in evaluation, demonstrating that failure coverage matters alongside semantic quality.

We introduce a failure-aware evaluation methodology and a reusable multi-dimension native-speaker rubric that future work can adopt. The key finding is not that romanized Indic is hard for LLMs (though it is), but that automatic metrics cannot be trusted as sole arbiters of quality for this task: evaluating romanized Indic generation requires a coordinated bundle of semantic similarity, script compliance, failure coverage, and native-speaker acceptability. As LLM deployment expands to non-English-speaking populations this gap becomes increasingly consequential---users typing in romanized script deserve the same quality assurance as those typing in English or native script, and this work provides a methodology to measure that quality and detect failures metrics alone cannot see.

\section{Limitations}
\subsection{Single Native Speaker Per Language}

This study uses one native speaker per language (Hindi, Telugu, Kannada). While this ensures language fluency and cultural authenticity, it prevents formal inter-rater agreement measurement. Future work with multiple raters per language would enable Cohen's $\kappa$ or Fleiss' $\kappa$ calculation.

\subsection{No Sampling Variance}

All models were evaluated at temperature 0, which ensures deterministic outputs. This prevents quantification of failure variance---we report point estimates without confidence intervals. A follow-up study with temperature $> 0$ would reveal whether specific prompts consistently trigger failures across multiple samples from the same model, or whether some failures are idiosyncratic to a single sample.

\subsection{Sample Size}

The diagnostic set (72 responses: 24 per language) is smaller than the full evaluation (1,704 responses), limiting the generalizability of human-evaluation findings. The diagnostic set was intentionally stratified on failure conditions to capture edge cases; it does not represent population failure rates. Population statistics (invalid row counts, BGE/BERTScore means) are computed over all 142 responses per model, but human evaluation covers a smaller diagnostic subset.

\subsection{Language Coverage}

We evaluate three Indic languages (Hindi, Telugu, Kannada). Other Indic languages (Tamil, Marathi, Bengali, Gujarati) and non-Indic romanized varieties (Arabizi, Finglish, Greek romanization) remain unexplored. Findings may not generalize beyond these three languages.

\subsection{Model Selection Bias}

We evaluate four models representing a mix of proprietary and open-source systems. Newer models or models specifically trained on Indic data (e.g., AI4Bharat models) were not tested and may show different failure profiles.

\subsection{Reference Task Dependency}

Evaluation uses OpenAssistant English pairs as the reference. Different source datasets or task distributions might reveal different failure modes. The choice of English-to-romanized-Indic as the generation direction (rather than romanized-to-native-script or native-to-romanized) reflects real-world user needs but limits scope.

\subsection{Unexplored Directions}

Two extensions are left to future work. First, we do not test explicit code-switching (e.g., English--Hindi code-mixed queries within a single message), which may exhibit failure profiles distinct from the monolingual romanization studied here. Second, we do not break results down by prompt category (emotional, technical, narrative); a finer per-task analysis could reveal which task types most often trigger script spillage or instruction collapse. Newer models and additional Indic languages (Tamil, Marathi) and Indic-specific fine-tuning are further natural directions.

\bibliography{references}

\appendix

\section{Output Token Efficiency}

Table~\ref{tab:output-tokens} reports mean model-reported output tokens for each paired query-response generation, including both generated fields and JSON syntax.

\begin{table}[H]
\centering
\footnotesize
\setlength{\tabcolsep}{4pt}
\begin{tabular}{lrrrr}
\toprule
\textbf{Model} & \textbf{Hindi} & \textbf{Telugu} & \textbf{Kannada} & \textbf{Overall} \\
\midrule
Claude 4.8 & 1,059 & 1,037 & 1,107 & 1,068 \\
GPT-5.5 & 714 & 691 & 737 & 714 \\
Gemini-2.5 & 633 & 686 & 717 & 678 \\
Qwen-3.5 & 714 & 447 & 541 & 567 \\
\bottomrule
\end{tabular}
\caption{Mean output tokens per model and language. Qwen Kannada excludes 49 degenerate rows; including them raises the Kannada mean to 1,768.}
\label{tab:output-tokens}
\end{table}

Claude reports the highest stable token usage (1,068 average), followed by GPT-5.5 (714) and Gemini (678). These differences reflect each model's tokenizer and generation strategy, not direct cost (each provider uses different tokenization). Prior work~\cite{petrov2023tokenization} has shown that non-English inputs incur a tokenization penalty: non-English content requires more tokens than English-equivalent content in the same model. Qwen3.5-35B is substantially less stable: its low Telugu mean (447) is partly caused by incomplete outputs, and its Kannada mean excludes degenerate rows. When degenerate rows are included, Kannada's token usage jumps to 1,768, suggesting the model enters unstable states on Dravidian languages.

\section{Evaluation Rubric}
\label{sec:rubric}

Our 10-dimension native-speaker rubric assigns 0--12 points total. Evaluators assess each dimension independently and provide justification in an ``Evaluator Notes'' field.

\begin{table*}[t]
\centering
\small
\begin{tabular}{@{}p{3.4cm}p{10.5cm}c@{}}
\toprule
Dimension & Description & Points \\
\midrule
1. Response Completeness & Is the response non-empty and reasonably thorough? & 0--2 \\
2. Script Purity & Are all characters Latin? No native script leakage? & 0--2 \\
3. English-Romanized Balance & Is code-mixing appropriate? (Some is natural, too much sounds foreign) & 0--3 \\
4. Romanization Consistency & Are word-to-romanization mappings consistent? No ambiguity? & 0--3 \\
5. Semantic Faithfulness & Does meaning carry from English through romanization? & 0--3 \\
6. Naturalness \& Fluency & Does it read naturally or sound like word-by-word translation? & 0--3 \\
7. Grammar Correctness & Verb conjugations, agreement, case marking correct? & 0--2 \\
8. Word Choice Accuracy & Are word choices appropriate and register-consistent? & 0--2 \\
9. Register Appropriateness & Does tone match the prompt? & 0--2 \\
10. Overall Usability & Would the evaluator use this response in real conversation? & 0--3 \\
\bottomrule
\end{tabular}
\caption{Native-speaker evaluation rubric (maximum 12 points total).}
\end{table*}

\subsection{Calibration}

A single LLM-generated reference score (Claude Opus 4.8) was provided as a calibration anchor only. Evaluators scored every response from scratch without reference to the model score, ensuring native-speaker judgment was independent.

\subsection{Scoring Examples}

\textbf{High-quality response (11--12 points):}
\begin{itemize}
    \item Natural romanization with no script spillage
    \item Appropriate minimal code-mixing
    \item Sounds composed in romanized Indic, not translated from English
    \item Register matches prompt (casual for chat, formal for advice)
\end{itemize}

\textbf{Acceptable response (8--10 points):}
\begin{itemize}
    \item Romanization correct but some translation artifacts
    \item Slight ambiguity or inconsistency in transliteration
    \item Semantic content preserved
    \item Evaluator would use it with minor edits
\end{itemize}

\textbf{Failing response (0--7 points):}
\begin{itemize}
    \item Script spillage, instruction collapse, or degenerate output
    \item Meaning lost or register inappropriate
    \item Evaluator would not use this response
\end{itemize}

\section{Failure Examples}
\label{sec:examples}

\subsection{Gemini 2.5-flash Script Spillage Examples}

\textbf{Telugu Row 53 (unexpected Russian):}
\begin{quote}
English prompt: ``How do you maintain work-life balance?''

Romanized response: ``Work-life balance is the only [{\cyrillicfont единственный}] key to happiness...''

(Russian word ``{\cyrillicfont единственный}'' \textit{yedinstvenniy}, ``the only one'' inserted mid-sentence)
\end{quote}
BGE-M3 score: 0.81 | Native score: 3/12

\textbf{Kannada Row 33 (wrong script contamination):}
\begin{quote}
Prompt: Ask for help with emotional decision

Response contains: Telugu script {\telugu అవును} (romanized: \textit{avuunu}, ``yes'') instead of Kannada romanization {\kannada ಹೌದು} \textit{haudu}
\end{quote}
BGE-M3 score: 0.82 | Native score: 2/12

\subsection{Qwen3.5-35B Instruction Collapse Examples}

\textbf{Telugu Row 45:}
\begin{quote}
English prompt: ``Write a romantic message to your friend in Telugu''

Qwen romanized response: ``Write a romantic message to your friend in Telugu''

(Exact English copy; instruction not followed)
\end{quote}
BGE-M3 score: 1.00 (perfect English-English similarity) | Native score: 0/12

\textbf{Kannada Row 67:}
\begin{quote}
English prompt: ``Kannada recipe: how to make idli''

Qwen romanized response: Returns English recipe unchanged
\end{quote}
BGE-M3 score: 0.98 | Native score: 0/12

\subsection{Claude Opus 4.8 High-Quality Example}

\textbf{Hindi Row 12:}
\begin{quote}
English prompt: ``How do you stay motivated when facing challenges?''

Claude romanized response: ``Jab aap mushkilon ka saamna karte hain, motivation badhane ke liye apne aap ko yaad rakhiye ki...''

Native script equivalent: {\devanagari जब आप मुश्किलों का सामना करते हैं, मोटिवेशन बढ़ाने के लिए अपने आप को याद रखिए कि...}

(Natural romanized Hindi with appropriate code-mixing; sounds composed, not translated)
\end{quote}
BGE-M3 score: 0.84 | Native score: 11/12

\end{document}